\documentclass[10pt,twocolumn,letterpaper]{article}

\usepackage{cvpr}              % To produce the CAMERA-READY version
\usepackage[dvipsnames]{xcolor}

\definecolor{cvprblue}{rgb}{0.21,0.49,0.74}
\usepackage[pagebackref,breaklinks,colorlinks,citecolor=cvprblue]{hyperref}
\usepackage[linesnumbered,ruled,vlined]{algorithm2e}
\usepackage{amsmath}
\usepackage{stmaryrd}
\usepackage{multirow}
\usepackage{colortbl}
\usepackage{xcolor}
\usepackage{graphicx}
\usepackage[linewidth=1 pt]{mdframed}
\definecolor{ppt:blue}{HTML}{0070C0}
\definecolor{ppt:red}{HTML}{E63946}
\definecolor{tab:blue}{HTML}{1f77b4}
\definecolor{tab:red}{HTML}{d62728}
\definecolor{alpha_green}{HTML}{C0E3C0}
\definecolor{alpha_red}{HTML}{d62728}
\definecolor{alpha_blue}{HTML}{1f77b4}
\SetCommentSty{textnormal}

\def\paperID{1833
} % *** Enter the Paper ID here
\def\confName{CVPR}
\def\confYear{2024}

\title{Ranking Distillation for Open-Ended Video Question Answering \\ with Insufficient Labels}

\author{Tianming Liang\textsuperscript{1}, Chaolei Tan\textsuperscript{1}, Beihao Xia\textsuperscript{2}, Wei-Shi Zheng\textsuperscript{1}, Jian-Fang Hu\textsuperscript{1}\thanks{Corresponding author.}\\
\textsuperscript{1}Sun Yat-sen University, China\\
\textsuperscript{2}Huazhong University of Science and Technology, China\\
{\tt\small \{liangtm, tanchlei\}@mail2.sysu.edu.cn, 
hujf5@mail.sysu.edu.cn
}
}

\begin{document}
\maketitle

\begin{abstract}
    This paper focuses on open-ended video question answering, which aims to find the correct answers from a large answer set in response to a video-related question.
    This is essentially a multi-label classification task, since a question may have multiple answers.
    However, due to annotation costs, the labels in existing benchmarks are always extremely insufficient, typically one answer per question.
    As a result, existing works tend to directly treat all the unlabeled answers as negative labels, leading to limited ability for generalization.
    In this work, we introduce a simple yet effective ranking distillation framework (RADI) to mitigate this problem without additional manual annotation. RADI employs a teacher model trained with incomplete labels to generate rankings for potential answers, which contain rich knowledge about label priority as well as label-associated visual cues, thereby enriching the insufficient labeling information.
    To avoid overconfidence in the imperfect teacher model, we further present two robust and parameter-free ranking distillation approaches: a pairwise approach which introduces adaptive soft margins to dynamically refine the optimization constraints on various pairwise rankings, and a listwise approach which adopts sampling-based partial listwise learning to resist the bias in teacher ranking.
    Extensive experiments on five popular benchmarks consistently show that both our pairwise and listwise RADIs outperform state-of-the-art methods. Further  analysis demonstrates the effectiveness of our methods on the insufficient labeling problem.
\end{abstract}

\vspace{-2em}
\section{Introduction}
\label{sec:intro}
Video question answering~\cite{zhong2022video,patel2021recent,le2020hierarchical} is one of the most popular research domains to explore the capability of AI models in understanding videos and languages. In this work, we rethink a fundamental task within this domain --- \textbf{open-ended video question answering (OE-VQA)}~\cite{xu2017video,yang2021justask,jang2017tgif,yu2019activitynet}, which requires the model to find the correct answers in a large vocabulary (with over 1K possible answers) in response to the question regarding a given video.

\begin{figure}[t]
    \centering
    \includegraphics[scale=0.4]{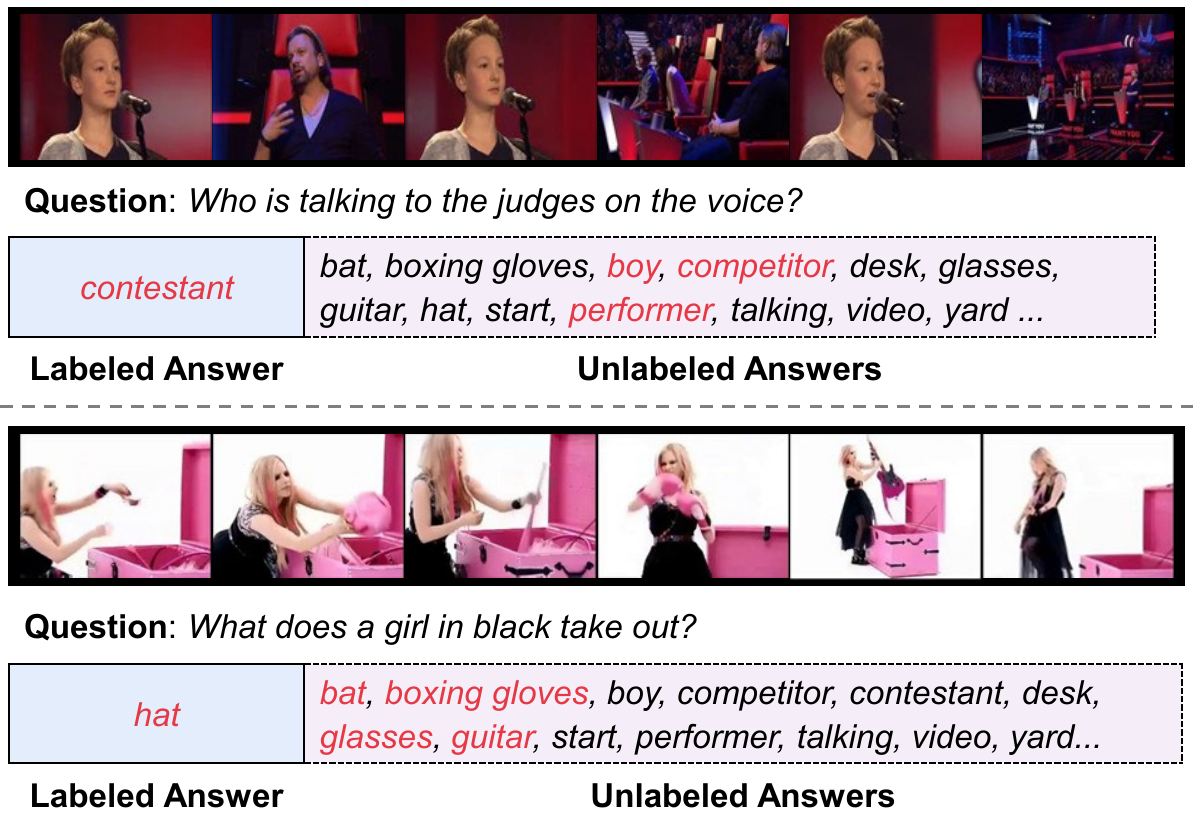}\vspace{-1em}
    \caption{Two examples about the insufficient labeling problem in MSRVTT-QA dataset. 
    % The first example ignores the labeled answer's synonyms, while the second example inherently has multiple answers.
    The correct answers to each questions are colored in \textcolor{ppt:red}{\textit{red}}.
    Existing OE-VQA methods tend to directly regard the entire unlabeled set as negative answers.
    }
    \label{fig:intro}\vspace{-1.5em}
 \end{figure}

Essentially, OE-VQA is supposed to be a multi-label classification task, as there is always more than one answer corresponding to a question. However, due to the cost of annotation, the labels provided by existing public benchmarks are extremely insufficient, typically one labeled answer per question, as illustrated in Figure~\ref{fig:intro}. 
In this study, we term this challenge as an \textbf{insufficient labeling  problem} in OE-VQA. 
This problem is critical but surprisingly neglected by even state-of-the-art approaches~\cite{yang2022frozenbilm,xiao2022video,li2023svitt,wang2023all}, which tend to formulate OE-VQA as a regular multi-class classification task by directly recognizing all the unlabeled answers as negative labels, as shown in Figure~\ref{fig:compare}(a).
These approaches ignore the numerous potentially correct answers, leading to a limited ability in open-world video question answering. 
To overcome this problem without extra manual labeling, three baseline schemes could be preliminarily considered:

\begin{figure}[t]
    \centering
    \includegraphics[width=1\linewidth]{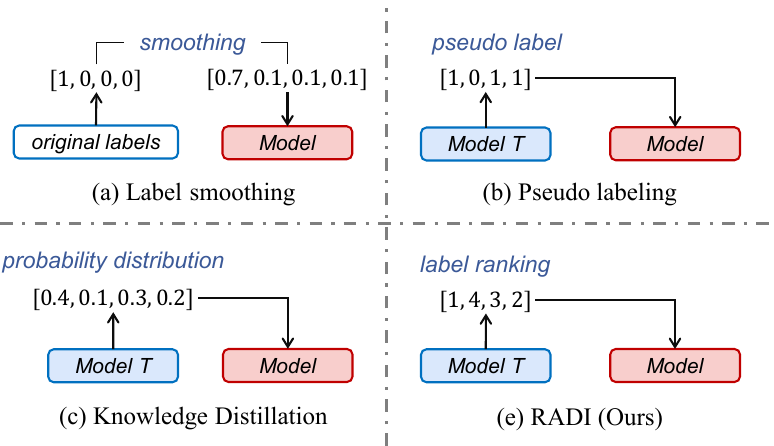}\vspace{-.8em}
    \caption{Comparison between different potential schemes for the insufficient labeling problem, where \textit{Model T} serves as a teacher model to enrich the label information.}
    \label{fig:compare}\vspace{-1.8em}
 \end{figure}

\textbf{(i)} \textit{Label smoothing} prevents overfitting the hard labels by replacing them with smoothed ones, as shown in Figure~\ref{fig:compare}(a). However, it treats all negative labels equally, failing to offer further hints about the potential answers.

\textbf{(ii)} \textit{Pseudo labeling} aims to generate pseudo labels with a trained model for label augmentation, as shown in Figure~\ref{fig:compare}(b). However, the accumulation of false positive labels during this process leads to incremental noise, thus limiting further improvements in performance.

\textbf{(iii)} \textit{Distribution distillation} aims to utilize the probability distributions of a trained teacher model to provide additional soft label information, as shown in Figure~\ref{fig:compare}(c). 
However, in the context of OE-VQA, only incomplete labels are available for training the teacher model. 
Distribution distillation is sensitive to the noise and bias in the imperfect teacher model, which limits its potential performance.

% This could introduce significant bias and noise into the teacher model, particularly in the context of OE-VQA, where the labels are extremely insufficient.

This work presents a \textbf{RAnking DIstillation framework} (\textbf{RADI}), which further mitigates the insufficient labeling problem by overcoming the limitations of the above baselines. 
As shown in Figure~\ref{fig:compare}(d), RADI utilizes answer rankings from a teacher model, which is trained with incomplete labels, to enrich the label information for training the student model.
Indeed, RADI can be regarded as a slack form of distribution distillation, since it relaxes the matching constraint from probability distribution to the label ranking.
In contrast to label smoothing and pseudo labeling, RADI can provide rich inter-label information while avoiding potential risks of hard pseudo labels. 
This is particularly beneficial for OE-VQA that typically involves a large label set.
Compared with distribution distillation, RADI is more robust to the insufficient labeling problem, since it depends on the relative ordering of answers instead of absolute scores.

% Both the teacher models in RADI and KD are noisy, what is the difference between them. We believe that in the absence of additional annotations, the noise and bias of the teacher model are difficult to avoid. Therefore, from distributed distillation to sequential distillation and then to the two methods proposed by us, all aim to gradually improve the robustness of distillation in noisy scenarios, rather than directly eliminate the noise in the teacher's model.

% This work presents a \textbf{RAnking DIstillation framework} (\textbf{RADI}) for OE-VQA, which overcomes the limitations of the above schemes and addresses the insufficient labeling problem effectively. Within RADI, we formulate KD as a \textit{learning-to-rank} (LTR) paradigm, and apply LTR objective functions to encourage the student model to learn the answer ranking of the teacher model, rather than the probability distributions. 
% In general, RADI preserves rich information about unlabeled answers with the ranked list, while mitigating the adverse impact due to the inaccurate prediction of the teacher model, by relaxing the constraint of original KD that directly matches the probability distribution.

RADI is flexible and can be integrated with existing off-the-shelf learning-to-rank (LTR) methods. However, regular LTR methods might not be adequate to address the extremely insufficient labeling problem, since they are sensitive to the ranking position of each answer. 
Indeed, strictly aligning the rankings between the teacher and student models may lead to convergence similar to that of distribution distillation.
Therefore, to further enhance the robustness of RADI, we design two alternate distillation approaches---\textit{pairwise ranking distillation} and \textit{listwise ranking distillation}.
In general, the pairwise approach tends to learn the relative priority by pairwise comparison, while the listwise approach aims to directly learn the absolute orders of a ranked list.
% Given a sorted list of $(a,b,c)$, the pairwise ranking approach tends to learn the ranking by pairwise comparison, such as $(a \rightarrow b)$, $(a \rightarrow c)$ and $(b \rightarrow c)$\footnote{In this paper, we use $(a \rightarrow b)$ to indicate $a$ ranks higher than $b$.}, while the listwise ranking approach aims to directly learn the global ranking, \ie, $(a \rightarrow b \rightarrow c)$. 
In the pairwise ranking distillation, we introduce soft margins to adaptively relax the optimization constraints to the uncertain pairwise rankings.
In the listwise ranking distillation, we design two ranking-based sampling strategies to enable distillation on partial list.
We empirically demonstrate that the two approaches can effectively avoid overconfidence in the teacher’s noisy rankings.
Moreover, both approaches are parameter-free, and bring no additional burden at inference.
We summarize our main contributions below:
\begin{itemize}
    \item We reveal the insufficient labeling problem in OE-VQA, and present an effective ranking distillation framework RADI to overcome it without extra manual annotation.
    \item We design two robust distillation strategies to further enhance the robustness of RADI to noisy rankings.
    \item We conduct extensive experiments on five popular OE-VQA benchmarks to demonstrate the improvement of RADI over state-of-the-art (SOTA) models and the effectiveness of the two distillation strategies.
\end{itemize}

% Although increasing the temperature in KD helps mitigate this to some extent, it tends to degrade KD to LS.

% In this work, we rethink the essential problem of OE-VQA.

\begin{figure*}[t]
    \centering
    \includegraphics[scale=0.45]{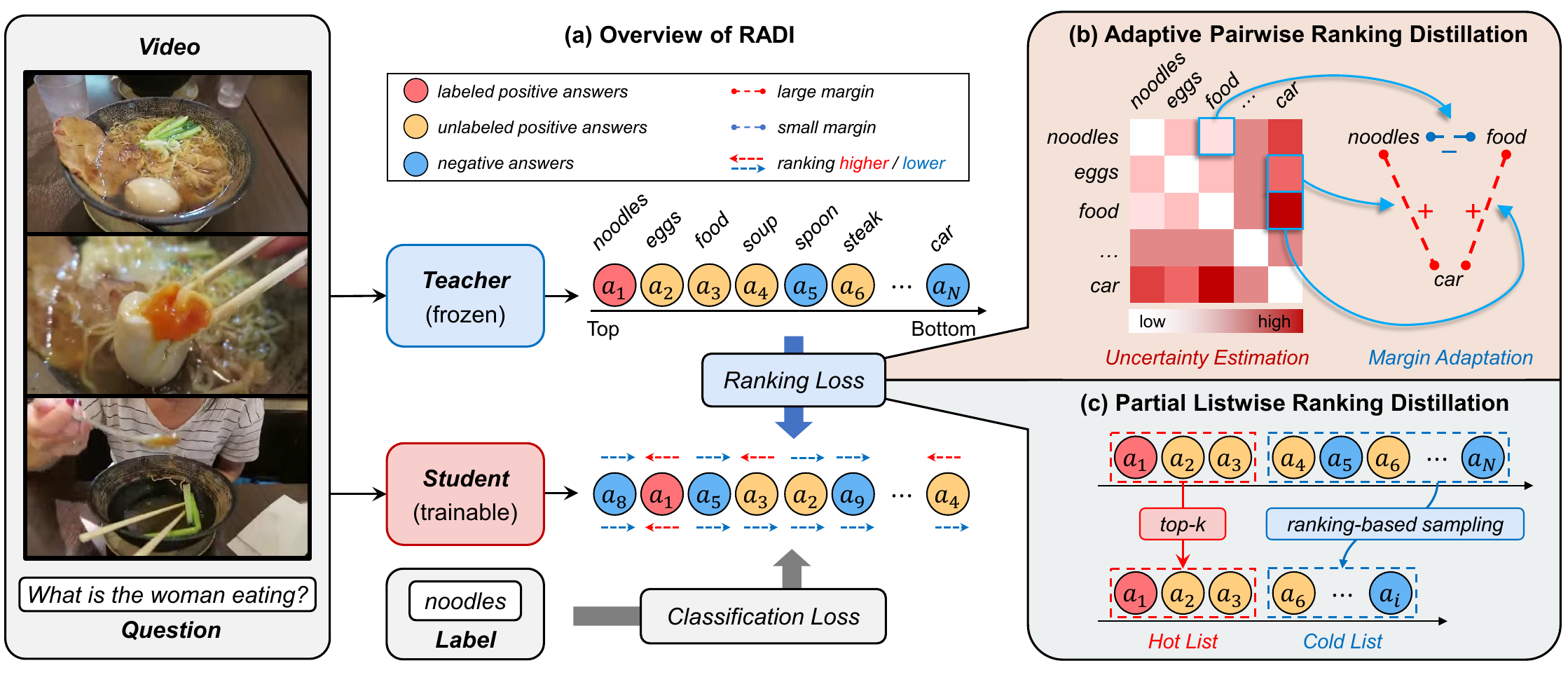}\vspace{-.5em}
    \caption{An overview of RADI, which is a LTR-based training framework for OE-VQA. Within RADI, the video-QA model is optimized using two loss functions: (i) \textbf{classification loss} maximizes the prediction probability of labeled answer (\ie, \textit{noodles} and \textit{eggs}) and suppress the rest, involving the potential positive answers (e.g., \textit{soup} and \textit{food});
    (ii) \textbf{ranking loss} may retrieve these potential positive answers, by pushing the predicted ranking to align with the ranking list provided by a well-trained teacher model. 
    }
    \label{fig:framework}\vspace{-1.5em}
 \end{figure*}

\section{Related Work}
\label{sec:related}
 
\noindent \textbf{Video question answering} aims to deduce an answer from a given video in response to a natural language question. There are mainly two types of tasks in this domain: the multiple-choice task \textit{MC-VQA}~\cite{lei2018tvqa,li2020hero} offers several options, typically up to five, for each question and requires selecting the single correct one, while the open-ended task \textit{OE-VQA}~\cite{xu2017video,yang2021justask,jang2017tgif,yu2019activitynet} provides a global vocabulary of over 1K possible answers and allows for multiple correct answers to a question. In this work, we mainly focus on OE-VQA because it is more challenging and practical~\cite{zhong2022video,patel2021recent}.
Existing efforts in OE-VQA are oriented towards two directions.
The first emphasizes the ability of \textit{understanding}, where the works aim to build strong video-question joint representations with memory networks~\cite{tapaswi2016movieqa,fan2019heterogeneous}, attention-based fusion modules~\cite{xu2017video,le2020hierarchical,lin2023collaborative}, or large-scale pretrained backbones~\cite{li2020hero,lei2021less,yang2021justask,wang2023all,yang2022frozenbilm}.
By contrast, the second direction emphasizes \textit{reasoning}, where the works explicitly model the objects and their interactions in videos for QA inference, by leveraging hierarchical structures~\cite{dang2021hierarchical,urooj2023learning,tan2023hierarchical} or graph neural networks~\cite{huang2020location,xiao2022video}.
% or symbolic program~\cite{yi2019clevrer,wu2021star,yi2018neural}.
Despite significant progress, the insufficient labeling problem, an essential limitation in OE-VQA benchmarks, is always ignored by these works. In this work, we formally reveal this problem and present promising non-manual solutions.

\noindent \textbf{Knowledge distillation} (KD)~\cite{hinton2015distilling} is a famous teacher-student learning paradigm in which a student model is trained by imitating the behavior of a trained teacher model. KD is widely applied in model compression~\cite{zhang2018shufflenet,kim2018paraphrasing,polino2018model,lin2022knowledge}, domain generalization~\cite{lin2022adversarial,lin2023diversifying} and transfer learning~\cite{xue2021multimodal,yang2021cross}.
% In model compression, the student model is typically much smaller than the teacher model, while in transfer learning, this is not a requirement.
% Hence, our work can be categorized into the latter field. 
According to the imitating objective, KD has three basic forms~\cite{gou2021knowledge}: 
\textit{distribution distillation}~\cite{hinton2015distilling,zhao2022decoupled,li2023curriculum} distills the output probability distributions, \textit{feature  distillation}~\cite{romero2014fitnets,zagoruyko2016paying,wang2019progressive} distills the intermediate features, and \textit{relation distillation}~\cite{yim2017gift,you2017learning,park2019relational} distills the feature relations between different layers or data samples.
The last two forms of KD emphasize feature learning, while distribution distillation focus more on label correlation learning. 
However, all these distillation methods could be suboptimal for the insufficient labeling problem, since they typically treat the teacher model as an infallible oracle, which is hard to achieve particularly in OE-VQA.
In this work, we overcome this limitation by introducing a slack ranking distillation framework RADI with two robust learning strategies, which distills the label information in a soft learning-to-rank manner. This manner reduces the sensitivity to the biased prediction and enhances the robustness to noisy knowledge of the imperfect teacher model.

\noindent \textbf{Learning to rank} is a task aiming to train a model for ranking a list of objects. In general, LTR models can be trained by \textit{pointwise}, \textit{pairwise} or \textit{listwise} methods. Pointwise methods~\cite{caruana1995using,crammer2001pranking,niu2016ordinal} treat each item individually with regression or classification. These method ignore the relationship among different items, typically leading to suboptimal performance. Therefore, recent works in LTR concentrate more on pairwise and listwise learning. Pairwise methods~\cite{burges2005learning,zhang2023led,wang2022sparsenerf} model the relative orders within individual pairs of items, while listwise methods~\cite{cao2007learning,xia2008listwise} directly model the absolute order of the entire list. 
% However, vanilla gradient-based listwise methods cannot directly maximize the non-smooth ranking metrics, e.g., Normalised Discounted Cumulative Gain (NDCG)~\cite{jarvelin2002cumulated}. Therefore, some works~\cite{burges2006learning,wang2018lambdaloss,zhu2020listwise,bruch2021alternative} try to establish metric-based listwise loss functions for better theoretical guarantees.
In this work, we propose a ranking distillation framework, which formulates the conventional teacher-student learning as a LTR paradigm. Furthermore, we propose two robust LTR approaches to mitigate the impact of the teacher model's bias in distillation.

% More recently, \citet{ko2023open} suggest extending OE-VQA to open-vocabulary evaluation. this is totally different to our paper in both motivation and techniques.
\section{Ranking Distillation for OE-VQA}
\label{sec:approach}
In this work, we propose a ranking distillation framework RADI to overcome the insufficient labeling problem in OE-VQA. As illustrated in Figure~\ref{fig:framework}(a), RADI employs a teacher model, which is merely trained with incomplete labels, to generate informative answer rankings, and then utilize these rankings as external labels to train the student model
 (Section \ref{subsec:overall}). To enhance the robustness of RADI to noisy rankings, we design two alternate distillation methods: \textit{adaptive pairwise ranking distillation}, as shown in Figure~\ref{fig:framework}(b), which adaptively adjusts the pairwise margins based on the teacher model's uncertainties about its pairwise rankings (Section \ref{subsec:pairwise});
and \textit{partial listwise ranking distillation}, as shown in Figure~\ref{fig:framework}(c), which performs partial listwise learning with a novel sampling strategy (Section \ref{subsec:listwise}).

\subsection{Overall Framework of RADI}
\label{subsec:overall}
\textbf{Task formulation.} Given an answer set $\mathbb{A}$, OE-VQA is formulated as a multi-label classification task, which requires the model to find the correct answers from $\mathbb{A}$ in response to a pair of video $v$ and question $q$.
Let $x$ denote a video-question pair, and $p_\theta(a_i|x)$ denote the prediction of a model $\theta$ for each individual answer $a_i$, then the normalized scores are achieved as follows:
\begin{equation}
    P_\theta(a_i|x) = \frac{\exp\left(p_\theta\left(a_i|x\right)\right)}{\sum_{j=1}^{N}\exp\left(p_\theta\left(a_j|x\right)\right)},
\end{equation}
where $N$ is the size of $\mathbb{A}$. 

\vspace{.5em}\noindent\textbf{Ranking distillation.} The training procedure of RADI consists of two stages. In the first stage, given a sample $x$ and the corresponding labeled set $\mathcal{A}_x$, we train a teacher model $\mathcal{T}$ with the classification loss $\mathcal{L}_{cls}$ as follows:
\begin{equation}
    \mathcal{L}_{cls}(\mathcal{T}|\mathcal{A}_x) = -\frac{1}{|\mathcal{A}_x|}\sum_{a \in \mathcal{A}_x} \log P_{\mathcal{T}} \left(a|x\right)
\end{equation}
Once the training of the teacher model is completed, we can obtain a ranked answer list $\mathcal{R}_x$ for each sample $x$, by sorting the teacher's predicted scores over all answers. In the second stage, we utilize both the original labels $\mathcal{A}_x$ and the ranking labels $\mathcal{R}_x$ to jointly train a student model $\mathcal{S}$:
\begin{equation}
    \mathcal{L}_x = \mathcal{L}_{cls}(\mathcal{S}|\mathcal{A}_x) + \alpha \mathcal{L}_{rank}(\mathcal{S}|\mathcal{R}_x),
\end{equation}
where $\mathcal{L}_{rank}$ is the ranking distillation loss that enables the student model to fit the teacher's answer ranking. 

Even with only the incomplete labels for training, the teacher model is still able to implicitly learn the label similarities as well as the associations between visual cues and labels owing to the extensive training data. 
% but it is suboptimal
This knowledge depicted in the teacher model can in turn enrich the original limited priori information, contributing to the training of the student model.
However, an inevitable challenge arises from the bias and noise inherent in the teacher's knowledge.
Directly using existing LTR methods for ranking learning is suboptimal, since they typically require perfect ranking labels.
To mitigate this challenge, we further design two robust distillation approaches in the subsequent sections.

\subsection{Adaptive Pairwise Ranking Distillation}
\label{subsec:pairwise}
The pairwise ranking approach aims to match the pairwise priorities between the teacher and the student models. This can be accomplished by the \textit{margin ranking loss} as follows:
\begin{equation}
    \setlength{\abovedisplayskip}{.5em}
    \mathcal{L}_p = \frac{1}{|\mathcal{R}_p|} \sum\limits_{(a_i, a_j)\in\mathcal{R}_p} \max \{0, m-\left(p_\mathcal{S}(a_i) - p_\mathcal{S}(a_j)\right)\}\label{eq:pair}
\end{equation}
where $\mathcal{R}_p = \{(a_i, a_j)|p_\mathcal{T}(a_i)>p_\mathcal{T}(a_j) \}$ is the set of positive pairwise priorities in the teacher model, $p_\mathcal{T}(\cdot)$ and $p_\mathcal{S}(\cdot)$ are non-normalized prediction probabilities of the teacher and student model, and $m$ is a constant hard margin. 

However, such hard-margin pairwise learning could be susceptible to noisy pairs, since it enforces the student model to match each pairwise priority from the imperfect teacher model without discrimination. 
To tackle this problem, we introduce adaptive soft margins that dynamically adjust the pairwise constraints based on the teacher model's uncertainty in its predictions.
% To tackle this problem, one can utilize the teacher model's beliefs in its predictions to reflect the reliability of each pairwise priority. 
For example, if the teacher model shows high uncertainty on a pairwise priority, then we should relax the learning for this pair, \ie, using a small margin. 
% Therefore, we expect our pairwise distillation be able to adaptively adjust the margins for different pairwise priorities based on the teacher model's uncertainties. 
% Actually, this purpose can be achieved by addressing the following two questions:
To this end, we propose soft pairwise ranking distillation, which enhances the robustness to noisy priorities by minimizing the disturbance of uncertain pairs.
Specifically, our approach is composed of two steps: the first step aims to estimates the teacher's pairwise uncertainties, and the second step optimizes the pairwise margins for minimizing the overall uncertainty.

\noindent\textbf{Uncertainty estimation.} Following recent works~\cite{cohen2021not,yang2022can} in uncertainty learning, we employ \textit{Monte Carlo Dropout}~\cite{gal2016dropout} to describe the uncertainty of the teacher model.
% \vspace{.5ex}
% \noindent (a) \textit{How to describe the teacher model's uncertainty on each pairwise priority?}
% \vspace{.5ex}
% Intuitively, if a model is highly uncertain about the priority between $a_i$ and $a_j$, then it may sometimes predict $a_i \rightarrow a_j$ and at other times $a_j \rightarrow a_i$. In other words, the prediction stability of a pair of answers can reflect the model's uncertainty about this pair. Inspired by this intuition, we introduce \textit{Monte Carlo Dropout}~\cite{gal2016dropout} to capture the uncertainty of the teacher model. 
In particular, for each sample, we forward the teacher model with dropout multiple times to derive multiple stochastic predictions. For the sake of efficiency, we only activate the dropout at the last layer, which implies the repetitive execution is only confined to the last layer. Let $p^k(a)$ denote the prediction score for an answer $a$ at $k$-th time, then we can compute the pairwise difference $\boldsymbol{D}^k_{ij} = p^k(a_i)-p^k(a_j)$ at each time. Finally, the uncertainty of the teacher model is represented by the variance of $\boldsymbol{D}$ over $T$ times as follows:
\begin{equation}
    \setlength{\abovedisplayskip}{.5em}\setlength{\belowdisplayskip}{.5em}
    \boldsymbol{U} = \frac{1}{T}\sum_{k=1}^T (\boldsymbol{D}^k - \bar{\boldsymbol{D}})^2,
\end{equation}
where \( \bar{\boldsymbol{D}} = \frac{1}{T}\sum_{k=1}^T \boldsymbol{D}^k \) is the mean matrix. We term $\boldsymbol{U}$ as the \textit{uncertainty matrix}, whose element $\boldsymbol{U}_{ij}$ quantifies the uncertainty of the teacher model on the corresponding pairwise priority $(a_i \rightarrow a_j)$. 

\setlength{\textfloatsep}{0.5em}
\begin{algorithm}[t]
    \caption{Sinkhorn algorithm for pairwise margin adaptation.}\label{alg1}
   \KwIn{
       Uncertainty matrix $\boldsymbol{U}$, smoothing factor $\lambda$. 
   }
   \KwOut{Margin scaling matrix $\boldsymbol{W}$.}
   
   Initialize $\boldsymbol{u},\boldsymbol{v} \gets \boldsymbol{1}/N$, $\boldsymbol{v}^{(0)} \gets \boldsymbol{1}$, $\Delta_v=1e^{8}$    \\
   $\widetilde{\boldsymbol{U}} \gets \exp\left(-\boldsymbol{U}/\lambda\right)$ \\
   $\operatorname{diag}(\widetilde{\boldsymbol{U}}) \gets 0$ \tcp*{Ignoring the self-pairs.}
   \While{$\Delta_v > 1e^{-8}$}
   {
       $\boldsymbol{u}^{(t)} \gets \boldsymbol{u} / (\widetilde{\boldsymbol{U}}\boldsymbol{v}^{(t-1)})$ \\
       $\boldsymbol{v}^{(t)} \gets \boldsymbol{v} /(\widetilde{\boldsymbol{U}}\boldsymbol{u}^{(t)})$ \\
        $\Delta_v \gets \operatorname{mean}(\lvert \boldsymbol{v}^{(t)} - \boldsymbol{v}^{(t-1)} \rvert)$ \\
    }
    $\boldsymbol{W}=\operatorname{diag}(\boldsymbol{u}^{(t)})\widetilde{\boldsymbol{U}}\operatorname{diag}(\boldsymbol{v}^{(t)})$
\end{algorithm}

\noindent\textbf{Margin adaptation.} The optimal margins of various pairs are determined by minimizing the overall uncertainty, since lower uncertainty tends to indicate higher reliability. This purpose can be achieved by formulating it as an \textit{Optimal Transport} (OT) problem~\cite{villani2009optimal,ge2021ota,chen2022plot}. 
Specifically, we regard each answer as a sender/receiver, and define the transmission cost from a sender $a_i$ to a receiver $a_j$ as their pairwise uncertainty $\boldsymbol{U}_{ij}$. In this way, our original target can be converted to a standard OT problem---\textit{finding the optimal transport plan with minimal total cost}, which can be efficiently addressed by off-the-shelf OT solvers, such as the Sinkhorn algorithm~\cite{cuturi2013sinkhorn}. We provide the pseudocode in Algorithm~\ref{alg1}. 
% Note that the output $\boldsymbol{W}$ is a distribution matrix. 
Finally, we replace the hard margin $m$ in \Cref{eq:pair} with the achieved soft margins 
$\boldsymbol{M}_{ij}=m\cdot\boldsymbol{W}_{ij}$.
% in OT, the sum of each row/column is 1, not the sum of the total matrix is 1.

% \vspace{.5ex}
% \noindent (b) \textit{How to adjust the margin for each pairwise priority in order to minimize the global uncertainty?}
% \vspace{.5ex}

\subsection{Partial Listwise Ranking Distillation}
\label{subsec:listwise}
The listwise ranking approach enables the model to directly learn the global ranking. As aforementioned, directly enforcing the student model to strictly match the rankings of the teacher model could be suboptimal, since the teacher model is trained with incomplete labels. 
To reduce the sensitivity to the biased rankings, we propose to distill only a partial ranking list rather than the entire. 
However, it is challenging to determine an appropriate ranking sublist for the partial distillation.
Indeed, this can be considered from two aspects.
On the one hand, the answers near the top of the entire list are significant because they get the highest attention from the teacher model. 
On the other hand, it is necessary to sample answers broadly from the entire list to support the global ranking. 
In light of these considerations, we propose a ranking-based sampling strategy, as depicted in Figure~\ref{fig:framework}(c). Firstly, we pick the top-$k$ answers from the teacher model's ranking list as the \textit{hot list}, and sample multiple answers from the remaining list to form the \textit{cold list}.
In building the cold list, we employ two alternative ranking-based sampling schemes: \textit{Exp-sampling} formulates the sampling probability of $k$-th answer as: $p_k \propto e^{-\alpha k}$, and \textit{Zipf-sampling} formulates the sampling probability as: $p_k \propto k^{-\alpha}$, where $\alpha$ is a smoothing coefficient.
Then, the hot list and the cold list are combined in sequence to derive the desired sublist. 
Within the sampled subset, we can perform listwise learning to encourage the student model to mimic the teacher model's partial rankings.

Our partial listwise ranking distillation is characterized by high degree of flexibility, enabling adaptation to any standard listwise loss functions, such as:

\noindent\textbf{ListMLE}~\cite{xia2008listwise} directly maximizes the likelihood of the target permutation $\mathcal{R}$ based on the Plackett–Luce model~\cite{plackett1975analysis}:
\begin{equation}
    \setlength{\abovedisplayskip}{.2em}\setlength{\belowdisplayskip}{.2em}
    \begin{aligned}
    \mathcal{L}_{\text{listmle}} &= -\log P(\mathcal{R}|\mathcal{S}) \\
    &= -\log \prod_{i=1}^n \frac{\exp \left(p_\mathcal{S}(a_{\mathcal{R}_i}) \right)}{\sum_{k=i}^n \exp \left(p_\mathcal{S}(a_{\mathcal{R}_k}) \right)},
    \end{aligned}
\end{equation}
where $n$ denotes the size of the sublist, and $a_{\mathcal{R}_i}$ denotes the answer ranked at $i$-th position in $\mathcal{R}$. 

\noindent\textbf{ListNet}~\cite{cao2007learning} considers the probability of each answer being ranked at the top (termed \textit{top-1 probability}), and thereby simplifying the lisewise learning into minimizing the cross entropy between the predicted scores and the target scores:
\begin{equation}
    \setlength{\abovedisplayskip}{.2em}\setlength{\belowdisplayskip}{.2em}
    \mathcal{L}_{\text{listnet}} = -\sum_{i=1}^{n} \phi\left(p_\mathcal{T}(a_i)\right) \log \phi\left(p_\mathcal{S}(a_i)\right),
\end{equation}
where $\phi(\cdot)$ is the \textit{softmax} function.

\noindent\textbf{STListNet}~\cite{bruch2020stochastic} extend ListNet from deterministic matching to stochastic matching by introducing random disturbance: 
\begin{equation}
    \setlength{\abovedisplayskip}{.2em}\setlength{\belowdisplayskip}{.2em}
    \mathcal{L}_{\text{stlistnet}} = -\sum_{i=1}^{n} \phi\left(p_\mathcal{T}(a_i)+\epsilon_i\right) \log \phi\left(p_\mathcal{S}(a_i)\right),
\end{equation}
where $\epsilon_i=-\beta\log(-\log u_i)$ and $u_i \sim \operatorname{Uniform}(0,1)$.

\noindent\textbf{LambdaLoss}~\cite{wang2018lambdaloss} is a family of metric-driven listwise loss functions, which can be uniformly defined as: 
\begin{equation}
    \setlength{\abovedisplayskip}{.2em}\setlength{\belowdisplayskip}{.2em}
    \mathcal{L}_{\text{lambda}} = -\sum_{(a_i,a_j)\in\mathcal{R}_p} \log_2 \left(\frac{1}{1+e^{-\sigma(p_\mathcal{S}(a_i)-p_\mathcal{S}(a_j))}}\right)^{w_{ij}},
\end{equation}
where $\sigma$ is a hyperparameter, and $w$ is called \textit{lambda weight}. Varying the definition of $w$ leads to various forms of LambdaLoss, such as \textbf{RankNet}~\cite{burges2006learning}, \textbf{NDCG-Loss1}, \textbf{NDCG-Loss2} and \textbf{NDCG-Loss2++}. Refer to~\cite{wang2018lambdaloss} for more details.

% with off-the-shelf listwise loss functions such as \textit{ListMLE}~\cite{xia2008listwise}, \textit{ListNet}~\cite{cao2007learning}, \textit{STListNet}~\cite{bruch2020stochastic} and \textit{LambdaLoss}~\cite{burges2006learning,wang2018lambdaloss}. \textcolor{red}{Details of these loss functions are presented in Supplementary.}

% Since the teacher model is frozen during distillation, the cold list would stay unchanged over iterations, while the hot list would change constantly due to the randomness in sampling. 

% \begin{equation}
%     p_k \propto e^{-\alpha k},
% \end{equation} 
% where $\alpha$ is a hyperparameter to control the smoothing of the distribution. The second scheme adopts the Zipf distribution, and defines the sampling probability as follows:
% \begin{equation}
%     p_k \propto \frac{k^{-\alpha}}{\sum_{k=c+1}^N k^{-\alpha}}.
% \end{equation}

% To keep the generality and flexibility of RADI, non-trivial sampling methods (e.g., selection networks~\cite{yang2019modeling,buch2022revisiting} and reinforcement learning~\cite{wu2019adaframe,chen2021decision}) are removed from our consideration. 

\section{Experiments}
\label{sec:exp}

\begin{table*}[t]
    \centering
    \small
    \begin{tabular}{lccccccccc}
        \toprule
        Model & \#Trainable Params & iVQA & ActivityNet-QA & MSVD-QA & MSRVTT-QA & TGIF-FrameQA \\
        \midrule
        SiaSamRea~\cite{yu2021learning} & - & - & 39.8 & 45.5 & 41.6 & 60.2 \\ 
        Just Ask~\cite{yang2021justask} & 157M & 35.4 & 39.0 & 47.5 & 41.8 & - \\
        MERLOT~\cite{zellers2021merlot} & 223M & - & 41.4 & - & 43.1 & 69.5 \\
        VIOLET~\cite{fu2021violet} & 198M & - & - & 47.9 & 43.9 & 68.9 \\
        Co-Token~\cite{piergiovanni2022video} & - & 38.2 & - & 48.6 & 45.7 & 62.5 \\
        SViTT~\cite{li2023svitt} & 255M & - & 43.2 & - & 43.0 & - \\
        All-in-one~\cite{wang2023all} & 110M & - & - & 48.3 & 46.8 & 66.3 \\
        FrozenBiLM~\cite{yang2022frozenbilm} & 30M & 39.6 & 43.2 & 54.8 & 47.0 & 68.6 \\
        \midrule
        \textbf{RADI-P (ours)} & 30M & \textbf{43.5} & \textbf{44.1} & \underline{55.8} & \textbf{48.2} & \underline{69.9} \\ 
        \textbf{RADI-L (ours)} & 30M & \underline{42.9} & \textbf{44.1} & \textbf{56.0} & \underline{48.1} & \textbf{70.0} \\ 
        \bottomrule
    \end{tabular}\vspace{-.5em}
    \caption{Comparison with SOTA models on multiple popular OE-VQA datasets. We use \textbf{RADI-P} and \textbf{RADI-L} to denote the RADI with pairwise ranking distillation and listwise ranking distillation, respectively.}
    \label{tab:sota}\vspace{-1.2em}
\end{table*}

In this section, we detail the experimental setup, and then conduct extensive experiments to demonstrate: (1) RADI achieves state-of-the-art performance on multiple mainstream OE-VQA datasets; (2) RADI outperforms other schemes on the insufficient labeling problem in OE-VQA. We also provide ablations to justify the design choices of our method, and present qualitative analyses to show the effectiveness of RADI.

\subsection{Experimental Setup}

\noindent\textbf{Datasets.} 
We select five famous OE-VQA datasets for evaluation: \textbf{iVQA}~\cite{yang2021justask}, {ActivityNet-QA}~\cite{yu2019activitynet}, \textbf{MSVD-QA}~\cite{xu2017video}, \textbf{MSRVTT-QA}~\cite{xu2017video} and \textbf{TGIF-FrameQA}~\cite{jang2017tgif}. 
% \textbf{iVQA}~\cite{yang2021justask} contains 10K video clips and 10K question, where each video-question pair is annotated with 5 overlapping answers. \textbf{ActivityNet-QA}~\cite{yu2019activitynet} contains 5.8K long-form videos with an average duration of 180s and 58K QA pairs. \textbf{MSVD-QA}~\cite{xu2017video} contains 1.8K video clips and 51K QA pairs. \textbf{MSRVTT-QA}~\cite{xu2017video} contains 10K video clips and 243K QA pairs. \textbf{TGIF-FrameQA} is the open-ended subset of TGIF-QA dataset~\cite{jang2017tgif}, which contains 46K GIFs and 53K QA pairs. 
% Among these datasets, only iVQA and ActivityNet-QA are manually annotated. 
Note that except for iVQA which annotates five ground-truth answers for each sample, all other datasets provide only one answer per sample. Therefore, we use iVQA as the main dataset for evaluation on the insufficient labeling problem.

\noindent\textbf{Metrics.} In this work, we are concerned about not only the correctness of the answer with the highest predicted score, but also the ranked positions of all correct answers. Hence for a comprehensive evaluation, we apply three distinct metrics: \textbf{Acc@1}, which indicates the top-1 accuracy; \textbf{Hit@5}, which checks whether at least one ground-truth answer appears within the top-5 prediction; and \textbf{nDCG@5}~\cite{jarvelin2002cumulated}, a popular ranking metric that measures the ranking quality at top-5 prediction by considering both the predicted positions and scores of the ground-truth answers. Note that in the comparison experiment with SOTA, we only apply Acc@1 due to the absent results for Hit@5 and nDCG@5 results in prior publications, while for the remaining experiments, we use all the three metrics on iVQA for evaluation.

% Before introducing nDCG, we need to define DCG@k as: 
% \begin{equation}
%     \operatorname{DCG}@k = \sum_{i=1}^k \frac{2^{y_i}-1}{\log_2(i+1)},
% \end{equation}
% where $y_i$ denote the ground truth score of the answer ranked at $i$-th position within the prediction list. Suppose \{\textit{car}, \textit{car}, \textit{taxi}, \textit{car}, \textit{car}\} is an annotated answer set in iVQA, and \{\textit{taxi}, \textit{bike}, \textit{car}\} is the top-3 prediction, then their ground truth scores are defined as their respective occurrence numbers in the annotated answer set, \ie, $\left(y_1, y_2, y_3\right) = \left(1, 0, 4\right)$. Finally, nDCG@k is obtained by:
% \begin{equation}
%     \operatorname{nDCG}@k = \frac{\operatorname{DCG}@k}{\operatorname{IDCG}@k},
% \end{equation}
% where IDCG@k is the ideal maximum value of DCG@k. In this example, it can be achieved when the top-2 prediction is \{\textit{car}, \textit{taxi}\}.
% nDCG@k is the most popular LTR metric, since it considers both the precisions and the ranks within the top-k prediction.

\noindent\textbf{Implementation details.} We use FrozenBiLM~\cite{yang2022frozenbilm} as the OE-VQA model for both the teacher and student. FrozenBiLM concatenates the frame embeddings and question embeddings in the sequential dimension as the input of a large pretrained language model DeBERTa~\cite{he2020deberta}, which then performs visual-textual joint interaction and outputs a matching score for each answer. During training, only a set of lightweight modules like adapters~\cite{houlsby2019parameter} and LayerNorm in FrozenBiLM are updated. For each video, we uniformly sample 10 to 20 frames, with the exact number varying across different datasets. For all datasets, we use AdamW as the optimizer, with a fixed learning rate of $5e^{-5}$ and the linear schedule with warmup. Following \cite{yang2022frozenbilm}, we use Dropout with probability of $0.1$ in adapters and gradient clipping with the maximum norm of $0.1$. In pairwise ranking distillation, we search for the margin scalar $m$ in $\{0.1, 1, 10\}$. In listwise ranking distillation, we normally fix the lengths of the cold list and the hot list as $10$ and $100$. 
% \textcolor{red}{The specific hyperparameter settings for different datasets are presented in Supplementary.}

\subsection{Comparison with State-of-the-arts}
We compare the top-1 accuracy of our RADI with SOTA models on five popular OE-VQA datasets. As shown in Table~\ref{tab:sota}, both the pairwise version RADI-P and the listwise version RADI-L consistently outperform the existing models on all the datasets. Especially on iVQA that provides multiple ground-truth answers, RADI-P and RADI-L improve the previous SOTA by 3.9\% and 3.3\% in terms of top-1 accuracy, respectively. This also suggests the effectiveness of our RADI on addressing OE-VQA with insufficient labels. Furthermore, both RADI-P and RADI-L maintain the same number of trainable parameter as FrozenBiLM, owing to our parameter-free pairwise and listwise LTR approaches.
In addition, although RADI-P and RADI-L learn to rank from different views, they achieve comparable performance on most of the datasets. This indicates both RADI-P and RADI-L are able to take full advantage of the label information in the ranking list.

\begin{table}[t]
    \small
    \centering
    \begin{tabular}{lccc} 
    \toprule
    Scheme & Acc@1 & Hit@5 & nDCG@5 \\
    \midrule
    FrozenBiLM & \text{40.5} & \text{64.5} & \text{49.6} \\
    \midrule
    \rowcolor{gray!15}  \textit{Label Smoothing} & ~ & ~ & ~ \\
    $\sigma=\text{0.1}$ & \text{40.2} \textcolor{tab:blue}{$\blacktriangledown_{\text{0.3}}$} & \text{65.0} \textcolor{tab:red}{$\blacktriangle_{\text{0.5}}$} & \text{49.5} \textcolor{tab:blue}{$\blacktriangledown_{\text{0.1}}$} \\
    $\sigma=\text{0.3}$ & \text{40.2} \textcolor{tab:blue}{$\blacktriangledown_{\text{0.3}}$} & \text{65.8} \textcolor{tab:red}{$\blacktriangle_{\text{1.3}}$} & \text{50.1} \textcolor{tab:red}{$\blacktriangle_{\text{0.5}}$} \\
    $\sigma=\text{0.5}$ & \text{38.7} \textcolor{tab:blue}{$\blacktriangledown_{\text{1.8}}$} & \text{64.4} \textcolor{tab:blue}{$\blacktriangledown_{\text{0.1}}$} & \text{48.7} \textcolor{tab:blue}{$\blacktriangledown_{\text{0.9}}$} \\
    \midrule
    \rowcolor{gray!15}  \textit{Pseudo Labeling} & ~ & ~ & ~ \\
    Top-\text{3} & \text{39.9} \textcolor{tab:blue}{$\blacktriangledown_{\text{0.6}}$} & \text{66.3} \textcolor{tab:red}{$\blacktriangle_{\text{1.8}}$} & \text{50.3} \textcolor{tab:red}{$\blacktriangle_{\text{0.7}}$} \\
    Top-\text{5} & \text{39.0} \textcolor{tab:blue}{$\blacktriangledown_{\text{1.5}}$} & \text{66.1} \textcolor{tab:red}{$\blacktriangle_{\text{1.6}}$} & \text{49.7} \textcolor{tab:red}{$\blacktriangle_{\text{0.1}}$} \\
    Top-\text{10} & \text{37.8} \textcolor{tab:blue}{$\blacktriangledown_{\text{2.7}}$} & \text{64.9} \textcolor{tab:red}{$\blacktriangle_{\text{0.4}}$} & \text{48.5} \textcolor{tab:blue}{$\blacktriangledown_{\text{1.1}}$} \\
    \midrule
    \rowcolor{gray!15} \multicolumn{2}{l}{\textit{Knowledge Distillation}} & ~ & ~ \\
    Vanilla KD~\cite{hinton2015distilling} & \text{41.3} \textcolor{tab:red}{$\blacktriangle_{\text{0.8}}$} & \text{66.7} \textcolor{tab:red}{$\blacktriangle_{\text{2.2}}$} & \text{51.2} \textcolor{tab:red}{$\blacktriangle_{\text{1.6}}$} \\
    FitNet~\cite{romero2014fitnets} & \text{39.2} \textcolor{tab:blue}{$\blacktriangledown_{\text{1.3}}$} & \text{63.2} \textcolor{tab:blue}{$\blacktriangledown_{\text{1.3}}$} & \text{48.2} \textcolor{tab:blue}{$\blacktriangledown_{\text{1.4}}$} \\
    RKD~\cite{park2019relational} & \text{39.2} \textcolor{tab:blue}{$\blacktriangledown_{\text{1.3}}$} & \text{63.8} \textcolor{tab:blue}{$\blacktriangledown_{\text{0.7}}$} & \text{48.6} \textcolor{tab:blue}{$\blacktriangledown_{\text{1.0}}$} \\
    DKD~\cite{zhao2022decoupled} & \text{41.4} \textcolor{tab:red}{$\blacktriangle_{\text{0.9}}$} & \text{67.4} \textcolor{tab:red}{$\blacktriangle_{\text{2.9}}$} & \text{51.5} \textcolor{tab:red}{$\blacktriangle_{\text{1.9}}$} \\
    CTKD~\cite{li2023curriculum} & \text{41.3} \textcolor{tab:red}{$\blacktriangle_{\text{0.8}}$} & \text{66.7} \textcolor{tab:red}{$\blacktriangle_{\text{2.2}}$} & \text{51.0} \textcolor{tab:red}{$\blacktriangle_{\text{1.4}}$} \\
    \midrule 
    \textbf{RADI-P (ours)} & \textbf{\text{43.2}} \textcolor{tab:red}{$\blacktriangle_{\text{\textbf{2.7}}}$} & \underline{\text{68.3}} \textcolor{tab:red}{$\blacktriangle_{\text{\underline{3.8}}}$} & \textbf{\text{52.7}} \textcolor{tab:red}{$\blacktriangle_{\text{\textbf{3.1}}}$} \\
    \textbf{RADI-L (ours)} & \underline{\text{42.9}} \textcolor{tab:red}{$\blacktriangle_{\text{\underline{2.4}}}$} & \textbf{\text{68.6}} \textcolor{tab:red}{$\blacktriangle_{\text{\textbf{4.1}}}$} & \textbf{\text{52.7}} \textcolor{tab:red}{$\blacktriangle_{\text{\textbf{3.1}}}$} \\
    \bottomrule
    \end{tabular}\vspace{-.5em}
    \caption{Evaluation on the insufficient labeling problem with iVQA. In label smoothing, $\sigma$ denotes the smoothing amount. In pseudo labeling, we employ the teacher model's Top-$k$ prediction as the pseudo labels, which are then combined with the original labels to train the student model. Note that the Acc@1 results here might slightly vary from that in \Cref{tab:sota}, since all schemes in this experiment can only access one labeled answer during training.}
    \label{tab:baseline}
\end{table}

\begin{figure*}[tb]
    \centering
    \begin{minipage}{0.335\linewidth}
        \centering
        \includegraphics[scale=0.32]
        {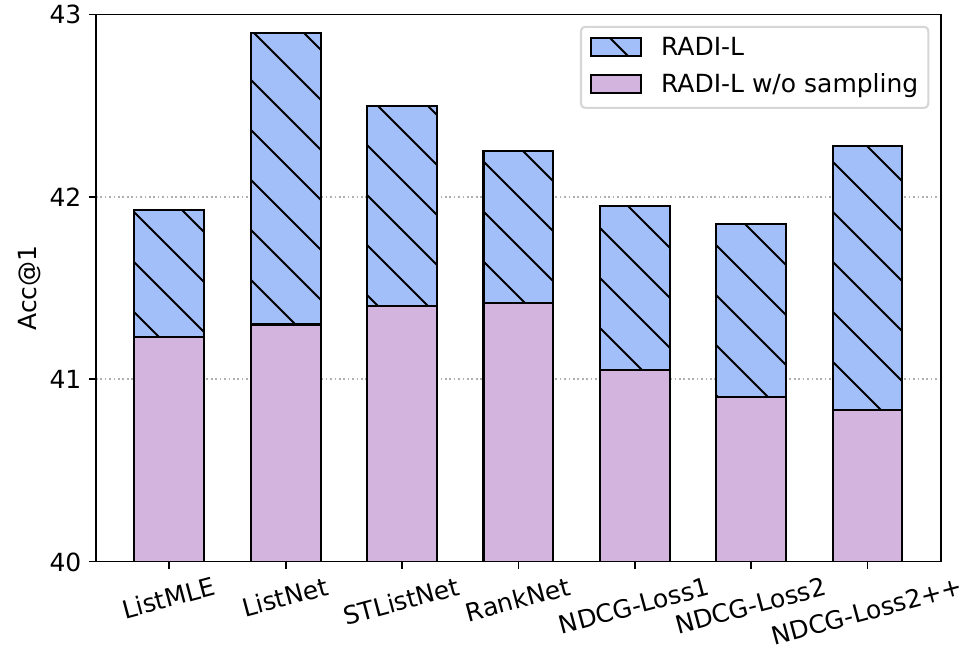}\vspace{-.5em}
        \caption{Improvements of using our sampling strategies on various listwise loss functions.}
        \label{fig:listwise}
    \end{minipage}\hfill
    \begin{minipage}{0.3\linewidth}
        \centering
        \includegraphics[scale=0.32]
        {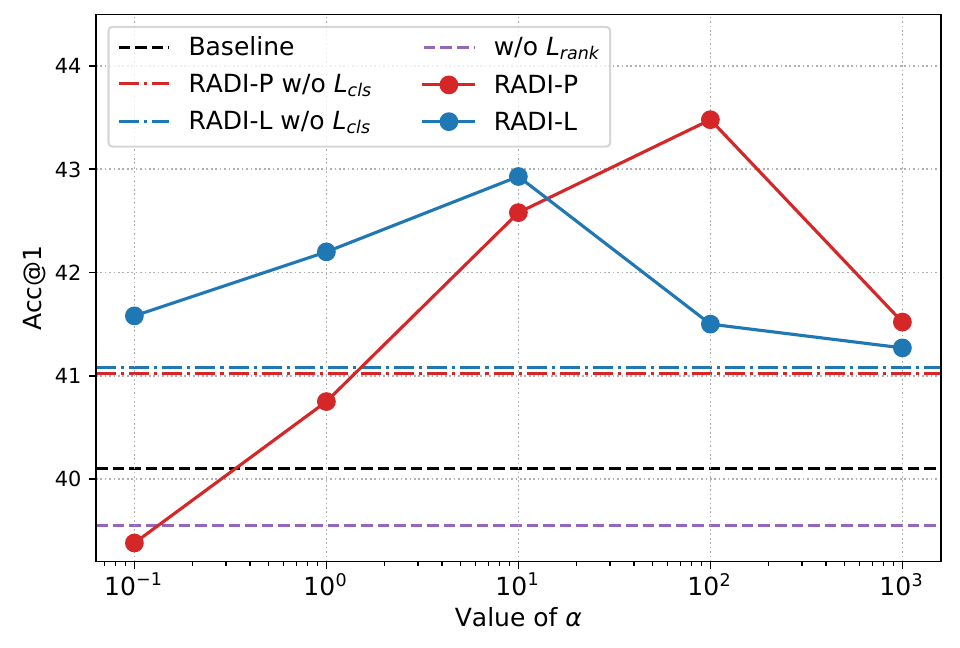}\vspace{-.5em}
        \caption{Impacts of using $L_{cls}$, $L_{rank}$ and different $\alpha$ on RADI-P and RADI-L.}
        \label{fig:alpha}
    \end{minipage}\hfill
    \begin{minipage}{0.3\linewidth}
        \centering
        \includegraphics[scale=0.32]
        {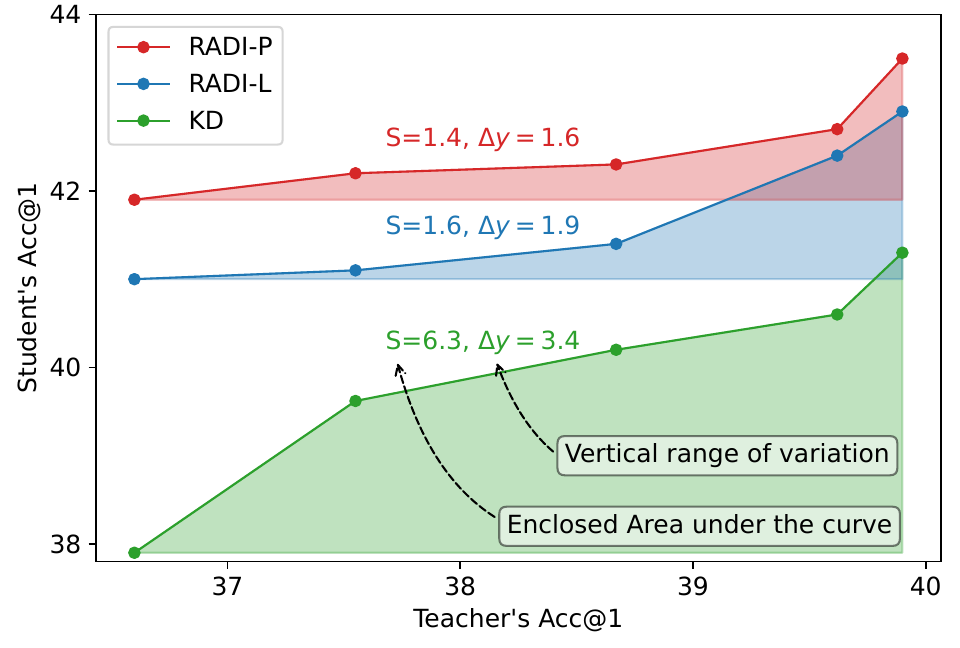}\vspace{-.5em}
        \caption{Impacts of using teacher models with different capacities.}
        \label{fig:teachers}
    \end{minipage}\vspace{-.7em}
\end{figure*}

\subsection{Evaluation on Insufficient Labeling Problem}
To demonstrate the effectiveness of RADI on the insufficient labeling problem in OE-VQA, we compare it with other potential schemes, as discussed in Section~\ref{sec:intro}, including \textit{label smoothing}, \textit{pseudo labeling} and various \textit{KD} methods on iVQA dataset in terms of Acc@1, Hit@5 and nDCG@5. For KD, we compare with a feature distillation method \textit{FitNet}~\cite{romero2014fitnets}, a relation distillation method \textit{RKD}~\cite{park2019relational}, and three distribution distillation methods: \textit{Vanilla KD}~\cite{hinton2015distilling} (which directly matches the distributions between teacher and student models), \textit{DKD}~\cite{zhao2022decoupled} (which decouples the distillation loss into target and non-target parts), and \textit{CTKD}~\cite{li2023curriculum} (which uses an adaptive temperature). 
As iVQA provides multiple annotated answers in both training and test sets, to better evaluate the insufficient labeling problem, we retain only the most frequently annotated answer for each training sample. In other words, all schemes \textbf{in this experiment} is trained with only one positive answer but evaluated with the complete positive answer set.
From the results shown in Table~\ref{tab:baseline}, we can observe that: 

\noindent\textbf{(i)} Slight label smoothing benefits to Hit@5, while heavy smoothing leads to significant performance degradation. In addition, label smoothing always results in a decline in Acc@1, since it tends to reduce the confidence of positive answers in exchange for confidence of the negative ones. 

\noindent\textbf{(ii)} Pseudo labeling performs better than label smoothing, but increasing the number pseudo labels leads to a rapid decline in Acc@1, due to the accumulation of noisy labels. 

\noindent\textbf{(iii)} The distribution distillation methods help to allevate the insufficient labeling problem, while the feature distillation method FitNet and the relation distillation method RKD consistently result in performance degradation. This shows the importance of the inter-label information introduced by distribution distillation, as mentioned in Section~\ref{sec:related}. 

\noindent\textbf{(iv)} Both our RADI-P and RADI-L show great superiority on the insufficient labeling problem, with significant improvements in all the metrics over the baseline schemes. In particular, we find RADI-P performs better in Acc@1 while RADI-L better in Hit@5, and both of them can achieve the best nDCG@5 results. These results demonstrate the effectiveness of our RADI and our improved LTR approaches.

\subsection{Ablation Study}
\noindent\textbf{Impact of pairwise ranking distillation.}
Our adaptive pairwise ranking distillation consists of two steps: uncertainty estimation and OT-based margin adaptation. Hence, we ablate them in this experiment. We design three pairwise baselines: \textit{Uniform} indicates all pairwise priorities share one margin $m$, which is the vanilla pairwise LTR method; \textit{Random} indicates the margin weight of each pairwise priority is a random sampled from $[0,1]$; and \textit{Uncertainty} indicates directly using the reciprocal of uncertainty as the margin weight. Here we term our pairwise strategy as \textit{Uncertainty+OT}. From the results shown in the upper part of Table~\ref{tab:ltr}, we can draw two observations: \underline{First}, both \textit{Uncertainty} and \textit{Uncertainty+OT} outperform \textit{Uniform} and \textit{Random}, which validates the effectiveness of our first step (\ie, uncertainty estimation). \underline{Second}, merely using \textit{Uncertainty} would lead to a decline in Acc@1, while \textit{Uncertainty+OT} benefits to all the metrics, which demonstrates the effectiveness of our second step (\ie, OT-based margin adaptation).

\begin{table}
    \footnotesize
    \centering
    \begin{tabular}{c|lccc} 
    \toprule
    \multicolumn{2}{c}{Strategy} & Acc@1 & Hit@5 & nDCG@5 \\
    \midrule
    \multirow{4}{*}{Pairwise} & \textit{Uniform} & 41.9 & 67.8 & 51.9 \\
    & \textit{Random} & 40.9 & 67.8 & 51.6 \\
    & \textit{Uncertainty} & 41.5 & 68.2 & 52.2 \\
    & + \textit{OT} & \textbf{43.5} & \textbf{68.3} & \textbf{52.7} \\
    \bottomrule
    \toprule
    \multirow{4}{*}{Listwise} & \textit{Full List} & 41.3 & 66.7 & 51.2 \\
    & \textit{Random} & 41.3 & 67.5 & 51.6 \\
    & \textit{Exponential} & 42.3 & 68.3 & 52.4 \\
    & \textit{Zipf} & \textbf{42.9} & \textbf{68.6} & \textbf{52.7} \\
    \bottomrule
    \end{tabular}\vspace{-.5em}
    \caption{Comparison of different strategies of pairwise ranking distillation and listwise ranking distillation on iVQA dataset.}
    \label{tab:ltr}
\end{table}

\noindent\textbf{Impact of listwise ranking distillation.}
In listwise learning, we propose to distill a partial list rather than the full list, and adopts two sampling strategies to achieve the partial list: \textit{Exp-sampling} and \textit{Zipf-sampling}. In this experiment, we compare them with two baselines: \textit{Full List} indicates distilling the full list, which is the vanilla listwise LTR method, and \textit{Random} indicates that the partial list is obtained by random sampling. All the methods in this experiment adopts ListNet~\cite{cao2007learning} as the loss function. As shown in the lower part of Table~\ref{tab:ltr}, the random sampling obtains similar results as \textit{Full List}, while both \textit{Exp-sampling} and \textit{Zipf-sampling} show significant improvements in all the metrics. In addition, we can also see on iVQA dataset, the employed \textit{Zipf-sampling} achieves the best performance. 

\noindent\textbf{Impacts of various LTR loss functions.}
As mentioned in \Cref{subsec:listwise}, our partial listwise ranking distillation can be integrated with multiple listwise loss functions. To demonstrate it, we present the results of using various listwise loss functions in Figure~\ref{fig:listwise}. It can be observed that our approach achieves consistent improvements across different loss functions, demonstrating the flexibility of our partial listwise ranking distillation.

\begin{table}[t] 
    \footnotesize
    \centering
    \vspace{-.5em}
    \setlength{\tabcolsep}{1.3mm}{
    \begin{tabular}{c|c|cccc}
      \toprule
      Dataset & \# \textit{answer} & Baseline & +KD & +RADI-L & +RADI-P \\
      \midrule
      iVQA & 2,349 & 6.5 & 8.0 & 8.0 & 8.2 \\
      TGIF-FrameQA & 911 & 8.5 & 11.1 & 11.1 & 11.5 \\
      ActivityNet-QA & 1,654 & 18.6 & 23.0 & 23.9 & 25.1 \\
      MSVD-QA & 1,198 & 28.5 & 36.3 & 36.3 & 36.6 \\
      MSRVTT-QA & 3,589 & 102.0 & 133.2 & 132.9 & 149.6 \\
      \bottomrule
    \end{tabular}}
    \vspace{-1em}
    \caption{Training time (\textit{minutes/epoch}) on a 3090 GPU. For fair comparison, the batch sizes are kept consistent.}\label{exp:time}
\end{table}

\begin{table}[t]
    \small
    \centering
    \begin{tabular}{lccc}
    \toprule
    Initialization & Acc@1 & Hit@5 & nDCG@5 \\ 
    \midrule
    \rowcolor{gray!15} \multicolumn{2}{l}{\textit{w/o Distillation}} & ~ & ~ \\
    From Scratch & 27.6 & 51.0 & 36.9 \\
    From Teacher & 40.0 & 66.5 & 50.3 \\
    Individual & 40.1 & 65.2 & 49.7 \\
    \midrule
    \rowcolor{alpha_red!15} \multicolumn{2}{l}{\textit{RADI-P}} & ~ & ~ \\
    From Scratch & 42.9 & 68.2 & 52.3 \\
    From Teacher & 42.8 & 67.9 & 52.3 \\
    Individual & 43.5 & 68.1 & 52.7 \\
    \midrule
    \rowcolor{alpha_blue!15} \multicolumn{2}{l}{\textit{RADI-L}} & ~ & ~ \\
    From Scratch & 42.3 & 68.7 & 52.6 \\
    From Teacher & 42.4 & 68.3 & 52.5 \\
    Individual & 42.9 & 68.6 & 52.7 \\
    \bottomrule
    \end{tabular}\vspace{-.5em}
    \caption{Results of different initialization for the student model.}
    \label{tab:students}
\end{table}

\noindent\textbf{Time cost.} Our RADI is an efficient parameter-free training paradigm, which incurs no extra burden at inference and only slightly increases the training time, as shown in Table~\ref{exp:time}. 
The efficiency stems from two aspects.
On one hand, RADI-L involves only two additional step---label ranking and sampling---over KD, thus the extra time cost is negligible.
On the other hand, RADI-P is efficient for that: 
1) Sinkhorn algorithm is an efficient approximation OT solver;
2) MC-dropout is applied to the last layer;
3) the pairwise matrices are truncated up to involving top 2500 answers. 

\noindent\textbf{Impacts of $L_{cls}$ and $L_{rank}$.}
In this experiment, we investigate the impacts of $L_{cls}$, $L_{rank}$, and using the different distillation coefficients $\alpha$ selected from ${0.1, 1, 10, 100, 1000}$. As shown in Figure~\ref{fig:alpha}, we can observe that: (i) continual training with only $L_{cls}$ leads to performance decline due to overfitting, while using only $L_{rank}$ can obtain further improvements; (ii) with the increase of $\alpha$, the benefits of RADI-L and RADI-P first increase and then decrease. Specifically, RADI-L performs better than RADI-P when $\alpha$ is small, and achieves the highest Acc@1 at $\alpha=10$. However, as $\alpha$ continues to rise, RADI-P shows further improvements and achieves the best at $\alpha=100$.

\noindent\textbf{Robustness to the imperfect teacher.}
We conduct ablation to validate the tolerance of our methods for imperfect teacher models. In this ablation, we evaluate the relation between distillation performance and the capacity of the teacher model, and present the results in Figure~\ref{fig:teachers}. As the teacher's performance changes, our RADI-P and RADI-L show more stable performance over KD. Specifically, when Acc@1 of the teacher models decreases by 3.3\%, the maximum Acc@1 decline of KD is 3.4\%, while that of RADI-P and RADI-L are merely 1.6\% and 1.9\%, respectively. More significantly, the area under the changing curve of KD is 4.5 and 3.9 times larger than that of RADI-P and RADI-L, respectively. These results demonstrate the benefits of our relaxation strategies to resisting the biased prediction of imperfect teacher models.

\noindent\textbf{Impact of the student initialization.}
There are three ways for initializing our student model: \textit{Scratch}, where the student model is initialized without pretraining on target datasets; \textit{Teacher}, where the student model is initialized with the weights of the trained teacher model; and \textit{Individual}, where the student model is pretrained on the target dataset in the same manner as the teacher model but with a different seed. As shown in Table~\ref{tab:students}, \textit{Individual} achieves the best performance for both RADI-L and RADI-P, hence we use it within RADI by default. In addition, it is noteworthy that the \textit{Scratch} initialization can still achieve competitive performance, which significantly shows the effectiveness and stability of our approach.

\begin{figure}[t]
    \includegraphics[scale=0.35]{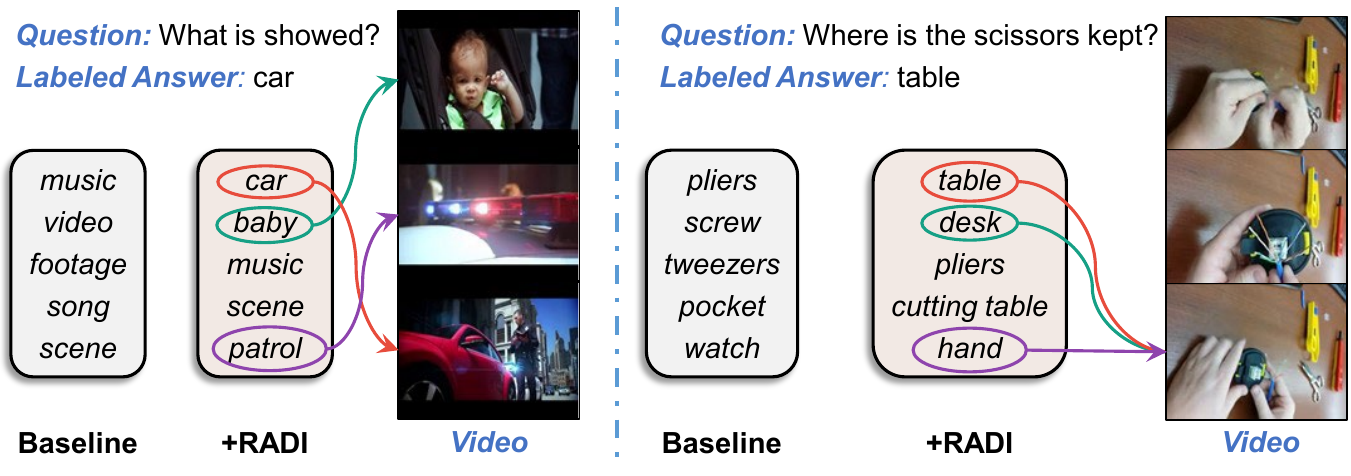}\vspace{-.5em}
    \caption{Qualitative results of the top-5 predictions. The arrows in colors connect the correct answers and the visual evidence.}
    \label{fig:visual}
 \end{figure}

\subsection{Qualitative Results}
We present the qualitative results in Figure~\ref{fig:visual} by comparing the top-5 predictions between the baseline model FrozenBiLM and our RADI. In the first example, there are at least three correct answers to the question according to the video, while only one answer "car" is labeled. However, RADI not only successfully predicts the labeled answer "car", but also finds two unlabeled correct answers "baby" and "patrol". In the second example, RADI correctly identifys the target position "table", and also finds the synonymous answers "desk" and "cutting table". These two cases show the effectiveness of RADI on the insufficient labeling problem.

\section{Conclusion}
In this paper, we focus on the insufficient labeling problem in OE-VQA task. To alleviate this problem without extra manual annotation, we present a simple but general ranking distillation framework RADI. 
It employs an imperfect teacher model trained with imcomplete labels to generate answer ranking for enriching the label information. 
To avoid overconfidence in the imperfect teacher model, we further design two robust and parameter-free ranking distillation approaches: adaptive pairwise ranking distillation with uncertainty-adaptive soft margins, and partial listwise ranking distillation with a ranking-based sampling strategy. We conduct extensive comparison and ablation experiments on five popular OE-VQA datasets to demonstrate the significant improvements of our pairwise and listwise RADIs.

{\footnotesize\vspace{.5em}\noindent\textbf{Acknowledgements}.
This work was supported partially by the NSFC (U21A20471, U22A2095, 62076260, 61772570), Guangdong Natural Science Funds Project (2020B1515120085, 2023B1515040025), Guangdong NSF for Distinguished Young Scholar (2022B1515020009), and Guangzhou Science and Technology Plan Project (202201011134).}

{
    \small
    \bibliographystyle{ieeenat_fullname}
    \bibliography{ref}
}



\end{document}